\documentclass[10pt,twocolumn,letterpaper]{article}

\usepackage[pagenumbers]{cvpr} % To force page numbers, e.g. for an arXiv version

\usepackage{xcolor}
\usepackage{tikz}
\usepackage{bbm}
\usetikzlibrary{fadings} % <-- ADD THIS LINE
\usetikzlibrary{shadings} % <-- THIS IS THE FIX. It enables color stops.
\usepackage{xspace}
\usepackage{wrapfig}
\usepackage[hypcap=true]{caption}
\definecolor{cvprblue}{rgb}{0.21,0.49,0.74}
\definecolor{TeacherColor}{RGB}{67, 170, 169}
\definecolor{StudentColor}{RGB}{157,122,217}
\definecolor{Sim2RealColor}{RGB}{124,170,51}

\definecolor{Blue80Cylan20}{RGB}{0,51,255}
\definecolor{Blue60Cylan40}{RGB}{0,102,255}
\definecolor{Blue40Cylan60}{RGB}{0,153,255}
\definecolor{Blue20Cylan80}{RGB}{0,204,255}

\usepackage[pagebackref,breaklinks,colorlinks,allcolors=cvprblue]{hyperref}

\def\paperID{15995} % *** Enter the Paper ID here
\def\confName{CVPR}
\def\confYear{2026}

\newcommand{\method}{{\texttt{VIRAL}}\xspace}

\title{\method: Visual Sim-to-Real at Scale for Humanoid Loco-Manipulation}

\author{Tairan He\textsuperscript{1,2*} \quad  Zi Wang\textsuperscript{1*}  \quad   Haoru Xue\textsuperscript{1,3*}   \quad   Qingwei Ben\textsuperscript{1,4*} \\ 
Zhengyi Luo\textsuperscript{1} \quad   Wenli Xiao\textsuperscript{1,2}  \quad  Ye Yuan\textsuperscript{1} \quad  Xingye Da\textsuperscript{1} \quad   Fernando Castañeda\textsuperscript{1}  \\
Shankar Sastry\textsuperscript{3}  \quad  Changliu Liu\textsuperscript{2} \quad   Guanya Shi\textsuperscript{2} \quad   Linxi ``Jim" Fan\textsuperscript{1\dag} \quad   Yuke Zhu\textsuperscript{1\dag}   \\
{\small \textsuperscript{1}NVIDIA  \quad \quad \textsuperscript{2}CMU  \quad  \quad \textsuperscript{3}UC Berkeley \quad  \quad \textsuperscript{4}CUHK \quad  \quad \textsuperscript{*}Equal Contribution \quad  \quad \textsuperscript{\dag}Project Leads}\\
\href{https://viral-humanoid.github.io}{\texttt{https://viral-humanoid.github.io}} 
}

\begin{document}
\twocolumn[{
\maketitle
\begin{center}
    \captionsetup{type=figure}
    \vspace{-25pt}
    \includegraphics[width=\linewidth]{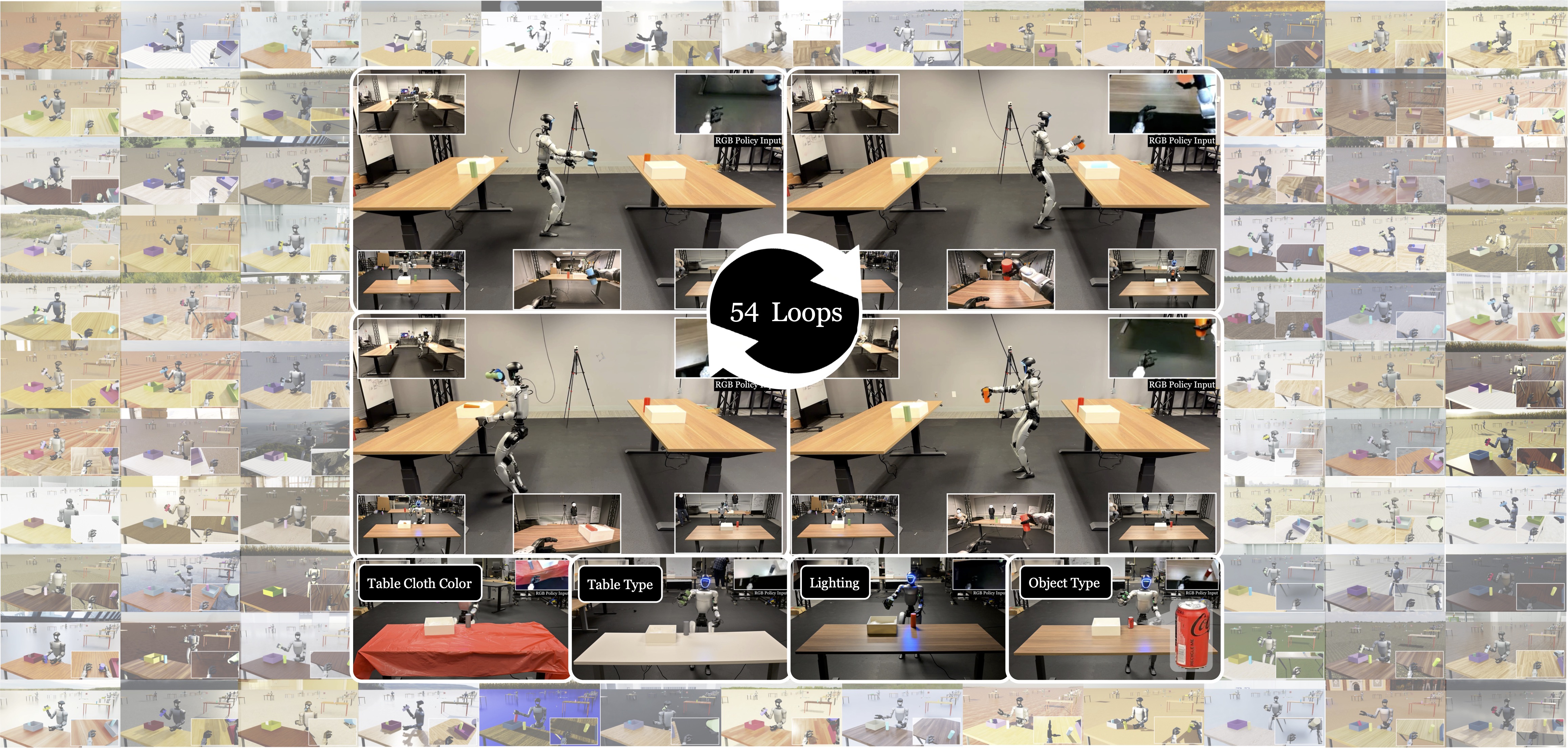}
    \vspace{-17pt}
\caption{\textbf{Center}: Unitree G1 humanoid performing loco-manipulation, walking between tables to place and pick objects for 54 loops with our RGB-based sim-to-real policy.
\textbf{Surrounding}: diverse simulated scenes used for training.
\textbf{Website}: \href{https://viral-humanoid.github.io}{\text{https://viral-humanoid.github.io}} 
}
\vspace{-5pt}
\end{center}
}]

\begin{abstract}
A key barrier to the real-world deployment of humanoid robots is the lack of autonomous loco-manipulation skills.
We introduce \method, a visual sim-to-real framework that learns humanoid loco-manipulation entirely in simulation and deploys it zero-shot to real hardware.
\method follows a teacher-student design: a privileged RL teacher, operating on full state, learns long-horizon loco-manipulation using a delta action space and reference state initialization.
A vision-based student policy is then distilled from the teacher via large-scale simulation with tiled rendering, trained with a mixture of online DAgger and behavior cloning.
We find that compute scale is critical: scaling simulation to tens of GPUs (up to 64) makes both teacher and student training reliable, while low-compute regimes often fail.
To bridge the sim-to-real gap, \method combines large-scale visual domain randomization over lighting, materials, camera parameters, image quality, and sensor delays—with real-to-sim alignment of the dexterous hands and cameras.
Deployed on a Unitree G1 humanoid, the resulting RGB-based policy performs continuous loco-manipulation for up to 54 cycles, generalizing to diverse spatial and appearance variations without any real-world fine-tuning, and approaching expert-level teleoperation performance.
Extensive ablations dissect the key design choices required to make RGB-based humanoid loco-manipulation work in practice.
\end{abstract}
\vspace{-15pt}

\vspace{-2pt}
\section{Introduction}
\label{sec:intro}
\begin{figure*}[tbp]
  \centering
  \includegraphics[width=1.0\linewidth]{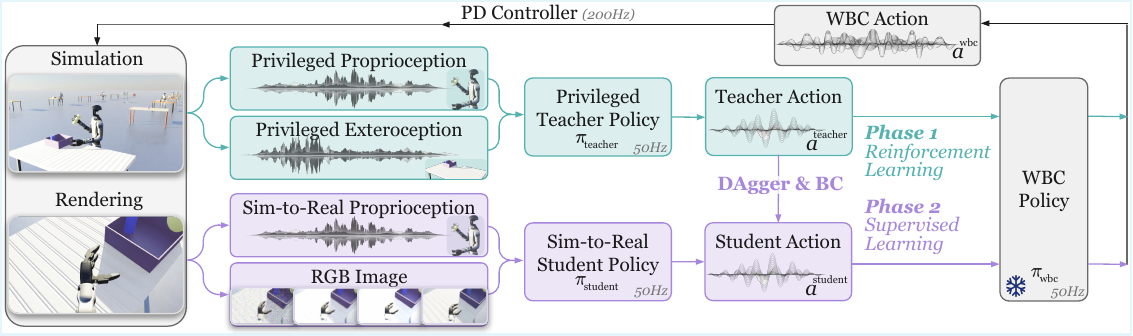}
  \vspace{-20pt}
  \caption{
  \textbf{\method teacher-student pipeline.}
  {\textcolor{TeacherColor}{Phase 1}}: In simulation, a privileged RL teacher policy $\pi_{\text{teacher}}$ receives full-state proprioception and exteroception of the task information and outputs WBC commands.
  {\textcolor{StudentColor}{Phase 2}}: A vision-based student policy $\pi_{\text{student}}$ observes only RGB images and sim-to-real proprioception and is trained to imitate the teacher policy via DAgger and behavior cloning.  }
  \label{fig:pipeline}
  \vspace{-14pt}
\end{figure*}

Humanoid robots are often framed as the natural embodiment of general-purpose physical intelligence: machines that could ultimately take on a large fraction of physical work for society.
Yet, despite rapid progress in hardware and control, current humanoids have delivered limited real, sustained productivity outside of carefully engineered demos~\cite{gu2025humanoid}.
A core missing piece is autonomous \emph{loco-manipulation}---tight coordination of locomotion and manipulation under onboard perception---over long horizons and across diverse environments to accomplish useful tasks.
Most existing humanoid systems either focus on blind locomotion~\cite{zhuang2024humanoid,liao2025beyondmimic,he2025hover,long2025learning}, static tabletop manipulation without mobility~\cite{lin2025sim,ze2024generalizable,maddukuri2025sim}, or rely heavily on human teleoperation~\cite{ben2025homie,zhang2025falcon,zhi2025learning,he2024omnih2o,ze2025twist2,luo2025sonic} or non-onboard sensors~\cite{weng2025hdmi,zhao2025resmimic}, and they rarely demonstrate autonomous loco-manipulation with onboard sensors in the real world~\cite{tirumala2024learning,dao2024sim,yin2025visualmimic}.

Recently, there has been an exciting push to replicate the large language model recipe~\cite{achiam2023gpt} in robotics, by collecting large-scale real-world datasets and training “robotic foundation models” from real-world teleoperation data~\cite{intelligence2504pi0,barreiros2025careful,o2024open,bjorck2025gr00t,zhao2024aloha,hu2024data,generalist2025gen0}.
While it remains unclear whether this path alone will suffice for general manipulation, it is clear that mobile manipulation will encounter substantially more variation than fixed tabletop setups and will therefore demand far more data~\cite{fu2024mobile,xiong2024adaptive,jiang2025behavior}.
When the mobile platform is a humanoid, the cost per data point increases even further due to hardware complexity, higher degrees of freedom, safety constraints, and the engineering overhead of the teleoperation stack~\cite{darvish2023teleoperation}.
In other words, if we treat humanoid mobile manipulation as “just another data problem,” the required scale may be prohibitively expensive in practice.

Simulation offers an alternative path.
Modern GPU-accelerated, photorealistic simulators can generate orders of magnitude more data at low marginal cost compared with human teleoperation~\cite{mittal2025isaac,makoviychuk2021isaac,authors2024genesis}.
Sim-to-real has become the de facto approach for legged locomotion~\cite{tan2018sim,rudin2022learning}, where policies trained in simulation routinely transfer to hardware~\cite{li2025reinforcement,cheng2023parkour,he2024agile}.
In contrast, manipulation is still largely dominated by imitation learning from real-world data, with sim-to-real successes typically restricted to tabletop settings and narrow tasks~\cite{akkaya2019solving,singh2024dextrah,handa2023dextreme,lin2025sim,chen2024object}.
Moreover, sim-to-real locomotion and manipulation are usually studied in isolation: locomotion work often ignores manipulation, and manipulation work typically assumes a fixed base.
In this paper, we aim to answer: \emph{Can visual sim-to-real enable useful humanoid loco-manipulation with onboard perception?}

Visual sim-to-real for robotics is not a new idea~\cite{zhu2018reinforcement,tirumala2024learning,liu2024visual,yuan2024learning,jiang2024transic,ze2023visual,hansen2021generalization,huang2022spectrum,akkaya2019solving,andrychowicz2020learning,singh2024dextrah,deng2025graspvla}, but we revisit it in the context of humanoid loco-manipulation and push the system to modern scales in simulation fidelity, GPU compute, and humanoid hardware.
Our goal is not to propose yet another novel RL or sim-to-real algorithm, but to provide a technical recipe on the full stack required to make RGB-based humanoid loco-manipulation work in practice: what designs matter, where they fail, and how they interact.

To enable efficient sim-to-real training, we adopt a teacher-student framework as shown in \Cref{fig:pipeline}. 
We first train an RL \emph{teacher policy} in simulation with full access to privileged state, operating on top of a pretrained whole-body control (WBC) policy~\cite{ben2025homie}. 
We then distill this teacher into a vision-based \emph{student policy} that observes only RGB images and proprioception accessible on the real robot. 
The student is trained with large-scale visual distillation using a mixture of online DAgger~\cite{ross2011dagger} and behavior cloning. We find that scaling up GPU compute for simulation training is essential for reliable learning of loco-manipulation skills.

To facilitate visual sim-to-real transfer, 
on the simulation side, we scale up visual randomization variations, including scene assets, lighting, materials, and camera parameters, with high-fidelity tiled rendering; on the hardware side, we align the simulator and real hardware to best match each other, including system identification (SysID) on high-gear-ratio dexterous hands and the alignment of cameras. 
Together, these technical elements yield an end-to-end RGB-based student policy that transfers zero-shot to the real humanoid robot and executes continuous loco-manipulation—walking, placing, grasping, and object transport—over long horizons. 

In real-world experiments, \method shows not only the robustness of the high success rate that is near the human expert teleoperation performance, but also generalization to various spatial and scene variations. 
In simulation experiments, scaling studies, and ablations reveal which key components of the \method framework are most critical for the full stack to work in practice. 
Overall, our results suggest that large-scale visual sim-to-real 
% with a structured teacher–student framework 
provides a practical path toward autonomous humanoid loco-manipulation.

\section{Key Elements of \method}
\label{sec:methods}

\noindent \textbf{\text{Framework Overview.}}
To achieve efficient visual simulation training, the \method controller is trained via teacher-student privileged learning~\cite{chen2020learning} as shown in \Cref{fig:pipeline}.
We first train a privileged RL \emph{teacher} policy with full access to the \textit{privileged state-based inputs} and run the \textit{simulation without the compute burden of visual rendering} on two 8-GPU L40S nodes (\textit{16 GPUs in total}). 
% \zi{Need to move training details to Experiment section.}
During this stage, we carefully design stage-based rewards and initialize environments from demonstrations to boost RL training. 
Instead of training low-level motor skills from scratch, we integrate the pre-trained WBC policy~\cite{ben2025homie} and make the command for WBC the action space for the teacher policy. Details of {\color{TeacherColor}teacher training} are provided in \Cref{Sec:teacher_elements}.

After the teacher discovers strong behavior under privileged information, we distill it into a \emph{student} policy that only receives the observations available on the real robot (\textit{i.e.,} proprioception and RGB images).
Visual distillation is performed using large-scale simulation on eight 8-GPU L40S nodes (\textit{64 GPUs in total}) with tiled rendering in Isaac Lab~\cite{mittal2025isaac}.
The student is trained by a combination of online DAgger~\cite{ross2011dagger} and behavior cloning to predict the teacher’s action given only access to proprioception and RGB image. More details of {\color{StudentColor}student training} are provided in \Cref{Sec:student_elements}.

To facilitate sim-to-real transfer of the RGB-based student policy, we randomize simulation assets, materials, dome lighting, image effects, camera extrinsics, and sensor delays during student training. We also perform real-to-sim alignment through system identification (SysID) of the Unitree 3-fingered dexterous hand and calibration of camera extrinsics. 
Finally, we deploy the student policy on the real robot without any fine-tuning.
Using onboard sensor observations, the student executes continuous loco-manipulation behaviors—including walking, placing, grasping, and object transport—on the Unitree G1 humanoid.
Details of {\color{Sim2RealColor}sim-to-real transfer} are provided in \Cref{Sec:sim2real_elements}.

\subsection{Key Elements of {\color{TeacherColor}Teacher} Training}
\label{Sec:teacher_elements}

{\color{TeacherColor}\textbf{\text{Teacher Formulation.}}}
We formulate the teacher as a goal-conditioned RL policy.
At time step $t$, the teacher $\pi_\text{teacher}(a_t | o_t^\text{priv})$ outputs a high-level command for the low-level WBC policy given privileged observation. Specifically, the teacher policy outputs $a_t = (\Delta \mathbf{v}_t,\, \Delta \boldsymbol\omega_t^{\text{yaw}}, \Delta \mathbf q^{\text{arm}}_t,\, \Delta \mathbf q^{\text{finger}}_t)$ as the command for the WBC policy~\cite{ben2025homie}, where $\Delta \mathbf{v}_t,\, \Delta \boldsymbol\omega_t^{\text{yaw}}$ are delta linear (x, y) and angular (yaw) velocity commands and $\Delta \boldsymbol q^{\text{arm}}_t,\, \Delta  \boldsymbol q^{\text{finger}}_t$ are delta joint targets for arm and finger motors.
These commands are passed to the WBC policy~\cite{ben2025homie}.
The privileged observation $o_t^\text{priv} = [o_t^\text{prop-priv}, o_t^\text{exte-priv}]$ includes privileged proprioception and exteroception. 
Proprioception consists of $o_t^\text{prop-priv} = [ \mathbf v_t, 
\boldsymbol\omega_t, 
\mathbf g_t,
\boldsymbol a_{t-1},
\boldsymbol q_t,
\boldsymbol{\dot{q}}_t,
\mathbf f^{\text{finger}}_t
]$
where
$\mathbf v_t,\boldsymbol\omega_t$ are base linear and angular velocities, 
$\mathbf g_t$ is base projected gravity, 
$\boldsymbol a_{t-1}$ is last action, 
$\boldsymbol q_t, \boldsymbol{\dot{q}}_t$ are joint positions and velocities, 
$\mathbf f^{\text{finger}}_t$ are fingertip forces. 
As for privileged exteroception, we have $o_t^\text{exte-priv} =[ e_t, \boldsymbol T_t, \boldsymbol O_t] $
where
$e_t$ is the current stage, 
$\boldsymbol T_t$ is the placement and lift target, 
$\boldsymbol O_t$ is the relative transforms of objects and tables to the robot. 
All observation terms are specified in \Cref{appendix:observation}.
The teacher is trained with PPO~\cite{schulman2017proximal} with a custom implementation of TRL~\cite{vonwerra2022trl} to train across GPUs in a distributed manner.

\noindent {\color{TeacherColor}\textbf{\text{Teacher Element \#1: Reward Design.}}} To design rewards for humanoid loco-manipulation, we segment the task into a sequence of walking, placing, grasping, and turning. Therefore, we define four key rewards: 
\begin{enumerate}
    \item \textit{Walking toward the objects}: $r_{\text{walk}}
    \;=\exp(-4\,(\|p_\text{robot} - p_\text{GraspObj}\| - 0.45)^2)$;
    \item \textit{Placing objects when near tray}: $r_{\text{place}}
    = - \|\mathbf f_{\text{PlaceObj}}\| * \mathbbm{1} (\|p_\text{PlaceObj} - p_\text{tray}\| < 0.3) $ where $\mathbf f_{\text{PlaceObj}}$ is the force between robot finger and the object to be placed;
    \item \textit{Grasping objects}: $r_\text{grasp-z} = \mathrm{min}(h_\text{GraspObj} - h_\text{table}, 0.15)$ 
    % \zi{We also have an upper bound for this reward} 
    and $r_\text{grasp-goal} = \exp(-10||p_\text{GraspObj} - p_\text{goal}||^2)$;
    \item \textit{Turning}: $r_\text{turn} = -|\text{y}_\text{robot} - \text{y}_\text{desired}|$ where y is the base yaw heading angle. 
\end{enumerate}
\noindent Full reward definitions are provided in \Cref{appendix:reward}.

\vspace{2pt}

% \subsection{Teacher - RL action space - Delta Action}
\noindent {\color{TeacherColor}\textbf{\text{Teacher Element \#2: Delta Action Space.}}}
Rather than outputting absolute joint targets—as is common in legged RL locomotion~\citep{rudin2022learning}—we adopt a \emph{delta} action space.
The policy outputs increments that are accumulated into the WBC command.
In practice, this delta representation significantly accelerates and stabilizes RL training.
An ablation of this choice is shown in \Cref{fig:TeacherResetDelta}.

\begin{figure*}[tbp]
  \centering
   \includegraphics[width=1.0\linewidth]{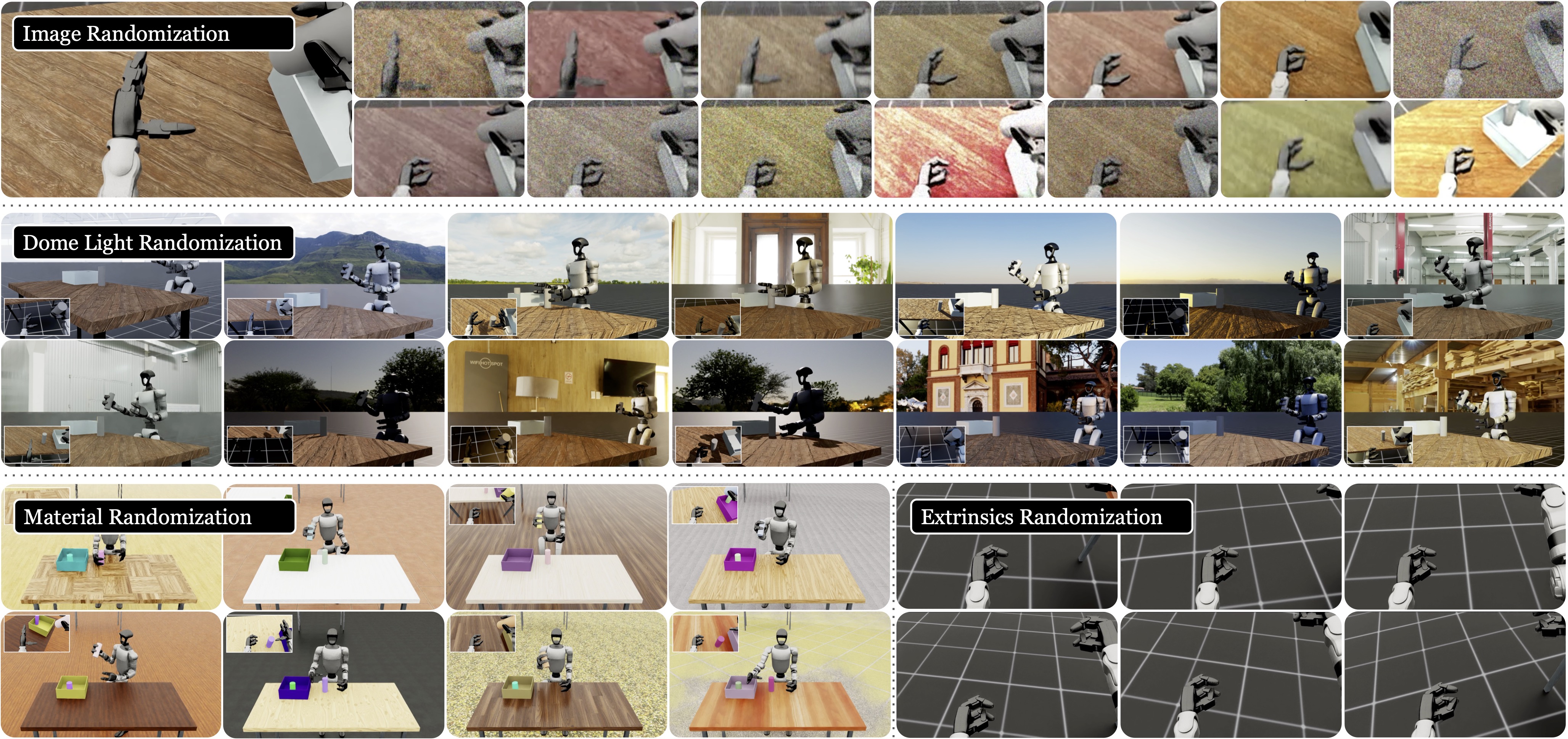}
   \vspace{-17pt}
   \caption{\textbf{Visual randomization} on
image, lighting, material, and camera-extrinsics randomization for sim-to-real robustness. 
}
\vspace{-10pt}
   \label{fig:Visual_Rand}
\end{figure*}

\vspace{2pt}

\noindent {\color{TeacherColor}\textbf{\text{Teacher Element \#3: WBC command as API.}}}
To reduce reward engineering burden and enable reliable real-world deployment, the teacher in \method produces high-level WBC commands rather than learning low-level motor skills from scratch.
We use HOMIE~\cite{ben2025homie} as the underlying WBC controller, which provides stable lower-body locomotion and diverse upper-body poses.
The command space of HOMIE involves velocity and height tracking commands for locomotion and upper-body joint commands. 
We extend this command interface by incorporating finger actions, yielding the full action space for \method.
Note that \method framework does not have designs overfitting to specific WBC policy, and can be extended to other humanoid WBC controllers~\cite{ze2025twist2,luo2025sonic}. With a stable and robust WBC policy as an API layer, the action space of \method policy is limited to a safe and reliable region of humanoid motions, improving deployability.

\begin{figure}[htbp]
  \centering
   \includegraphics[width=1.0\linewidth]{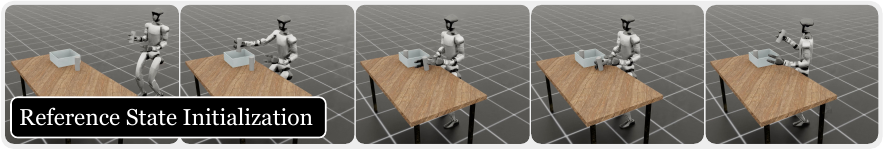}
    \vspace{-17pt}
   \caption{Frames of reference state initialization for teacher RL.}
   \vspace{-13pt}
   \label{fig:TeacherReset}
\end{figure}

\vspace{2pt}
% \subsection{Teacher - RL exploration - Demonstration Initilization}
\noindent {\color{TeacherColor}\textbf{\text{Teacher Element \#4: Reference State Initialization.}}} 
Learning long-horizon walking-placing-grasping-turning skills for high-DoF humanoids with RL typically demands heavy reward engineering still often yields suboptimal or poor sim-to-real transfer. 
To mitigate this, we collect 200 teleoperated \emph{simulation} demonstrations and use them as a state-initialization buffer for RL (\Cref{fig:TeacherReset}).
At every episode reset, we sample a demonstration snapshot and initialize the scene---robot, objects, and tables---accordingly, exposing the policy to diverse rewarding states long before it is capable of reaching them from scratch.
This reference-biased exploration greatly reduces reliance on brittle reward tuning and improves sim-to-real transfer, as human-provided grasping and placement poses offer strong priors.
Similar ideas have appeared in humanoid control~\cite{peng2018deepmimic,sharon2005synthesis} and manipulation~\cite{lin2025sim,nair2018overcoming}.
We find this reset strategy to be essential for training humanoid loco-manipulation, as shown in the ablation in \Cref{fig:TeacherResetDelta}.

\subsection{Key Elements of {\color{StudentColor}Student} Training}
\label{Sec:student_elements}

\noindent {\color{StudentColor}\textbf{Student Element \#1: DAgger\&BC Mixture.}}
We train the RGB-based student policy by distilling from the privileged teacher through a hybrid of online DAgger~\cite{ross2011dagger} and behavior cloning (BC). Both procedures share the same MSE objective, computed over a mixture of teacher- and student-induced observation distributions:
\begin{align*}
\rho^o &\triangleq \alpha\,\rho^o_{\pi_{\text{teacher}}}
+
(1-\alpha)\,\rho^o_{\pi_{\text{student}}}, \\ %\\[4pt]
\mathcal{L}_{\text{distill}}
&=
\mathbb{E}_{o_t\sim\rho^o}
\left[
\left\|
\pi_{\text{teacher}}(o^{\text{teacher}}_t)
-
\pi_{\text{student}}(o^{\text{student}}_t)
\right\|_2^2
\right],
\end{align*}
where $\rho^o_{\pi_{\text{teacher}}}$ and $\rho^o_{\pi_{\text{student}}}$ denote the observation distributions induced by the teacher and student rollouts, and $\alpha$ denotes the ratio. The distinction between DAgger and BC lies solely in the source of observations: teacher rollouts provide clean, near-optimal demonstrations that rapidly imprint strong priors on the student, while student rollouts expose the learner to states outside the teacher’s ideal distribution, which is critical for improving error-correction robustness and preventing compounding error during deployment. This mixed-policy rollout combines the fast initialization of BC with the state-coverage benefits of DAgger, producing a more resilient vision-based controller. Ablation of the mixture coefficient $\alpha$ is provided in \Cref{fig:teacher_rollout_ratio}.

\vspace{1mm}
\noindent {\color{StudentColor}\textbf{Student Element \#2: Network Backbone.}}
For the student's vision backbone, we adopt a state-of-the-art image encoder~\cite{simeoni2025dinov3} to extract high-quality RGB features, which are fused with proprioceptive to the policy head. 
The resulting student observation $o^{\text{student}}$ therefore integrates both visual embeddings and the proprioception available on real hardware, enabling the policy to reason over rich visual cues while maintaining grounded low-level awareness. We also evaluate choices for the student policy head, including a single-step MLP and a history-aware architecture that incorporates temporal context. 
Ablations of the vision backbone and history architecture are shown in \Cref{fig:VisionBackbone} and \Cref{fig:HistoryArch}, respectively.

\vspace{1mm}
\noindent {\color{StudentColor}\textbf{Student Element \#3: Distributed Simulation Learning System.}}
Large-scale visual simulation is substantially more expensive than rendering-free physics, typically operating at least an order of magnitude slower in terms of simulation throughput.
To scale up visual simulation training throughput, 
we implement a customized version of TRL~\cite{vonwerra2022trl} with support of Accelerate~\cite{accelerate} for efficient scaling across multiple GPUs and compute nodes. 
This implementation preserves the simplicity of single-GPU training while enabling near-linear scaling to large clusters for high-throughput visual sim-to-real learning. We identify scaling up GPUs for both teacher and student training as critical in our ablation studies in \Cref{fig:TeacherScaling} and \Cref{fig:StudentScaling}.

\subsection{Key Elements of {\color{Sim2RealColor}Sim-to-Real} Transfer}
\label{Sec:sim2real_elements}

\begin{figure}[htbp]
  \centering
  \vspace{-5pt}
   \includegraphics[width=1.0\linewidth]{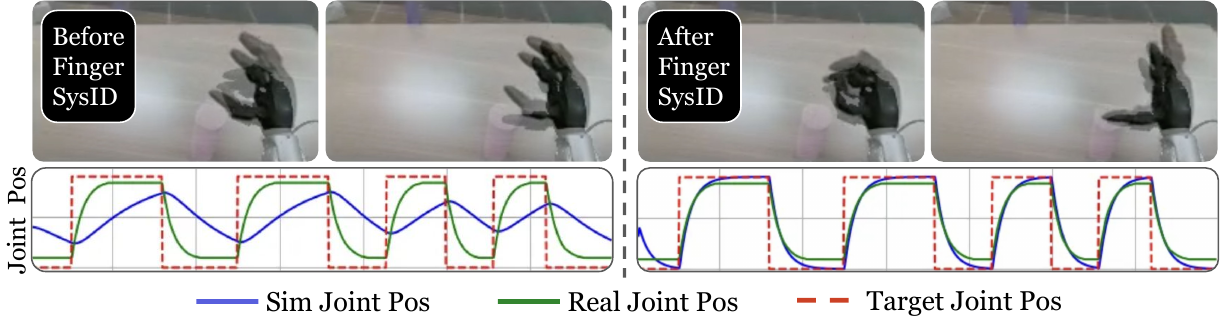}
    \vspace{-19pt}
   \caption{\textbf{System identification of the dexterous hand.} Real–sim overlays (top) and joint position trajectories (bottom) before and after SysID, showing markedly improved alignment.}
   \vspace{-12pt}
   \label{fig:Finger-SysID}
\end{figure}

\noindent {\color{Sim2RealColor}\textbf{Sim-to-Real Element \#1: SysID for Dexterous Hand.}}
While modern humanoids increasingly use low–gear ratio motors—reducing the need for motor-level SysID—the Unitree G1’s 3-fingered dexterous hand employs high gear ratios, resulting in a substantial sim-to-real mismatch. 
To address this, we define a real-world grasp–release primitive and replay the identical action sequence in simulation. 
We then perform SysID over finger armature, stiffness, and damping parameters to align simulated joint trajectories with real measurements. 
As shown in \Cref{fig:Finger-SysID}, SysID significantly improves the correspondence between simulated and real joint positions.

\begin{figure}[htbp]
  \centering
  \vspace{-5pt}
   \includegraphics[width=1.0\linewidth]{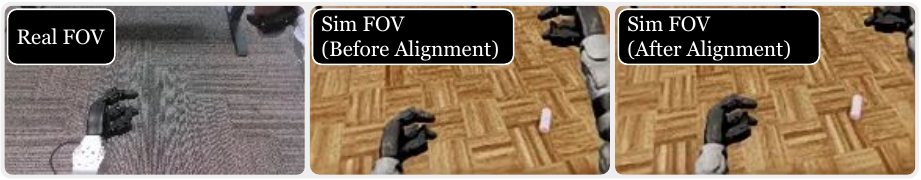}
   \vspace{-19pt}
   \caption{\textbf{Real-to-sim camera extrinsics alignment.}
Real view versus simulated views before and after alignment. 
}
   \label{fig:FOV-Alignment}
   \vspace{-5pt}
\end{figure}

\noindent {\color{Sim2RealColor}\textbf{Sim-to-Real Element \#2: FOV Alignment and Randomization.}}
We match the simulator’s camera intrinsics (focal length, focus distance, and sensor apertures) to the manufacturer's specifications. 
However, the camera extrinsics of Unitree G1 robots vary across units due to mechanical tolerances and can even drift over time on the same robot. 
To better align simulated and real visual observations, we perform a lightweight real-to-sim extrinsics calibration by visually matching rendered and real images (\Cref{fig:FOV-Alignment}). 
We further apply extrinsics randomization during training (\Cref{fig:Visual_Rand}) to ensure that the student remains robust to hardware-induced viewpoint differences.

\vspace{1mm}
\noindent {\color{Sim2RealColor}\textbf{Sim-to-Real Element \#3: Visual and Simulation Randomization.}}
To enhance robustness and improve sim-to-real transfer, we apply extensive visual and physical randomization during training (\Cref{fig:Visual_Rand}). 
We randomize image quality (brightness, contrast, hue, saturation, Gaussian noise, and blur), camera extrinsics to account for small pose shifts, and camera latency to model transmission delays. 
We additionally vary global illumination using dome-light environments and randomize material and color properties of floors, tables, objects, and robot components. 
These perturbations significantly improve the policy's transferability by preventing overfitting to any specific simulated appearance or lighting condition. Ablation of this randomization is provided in \Cref{fig:VisualRandomizationAbaltion}.

\begin{figure}[tbp]
  \centering
   \includegraphics[width=1.0\linewidth]{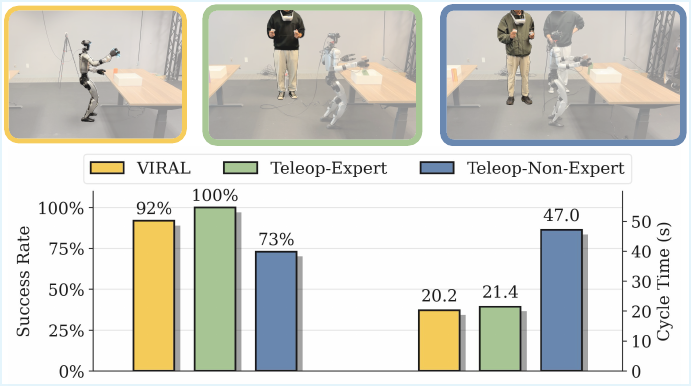}
   \vspace{-17pt}
   \caption{
\textbf{Real-world performance comparison:}
\method matches expert-level reliability, outperforms non-experts, and operates faster than the expert teleoperator.}
\vspace{-15pt}
   \label{fig:real_exp}
\end{figure}

\begin{figure*}[tbp]
  \centering
   \includegraphics[width=1.0\linewidth]{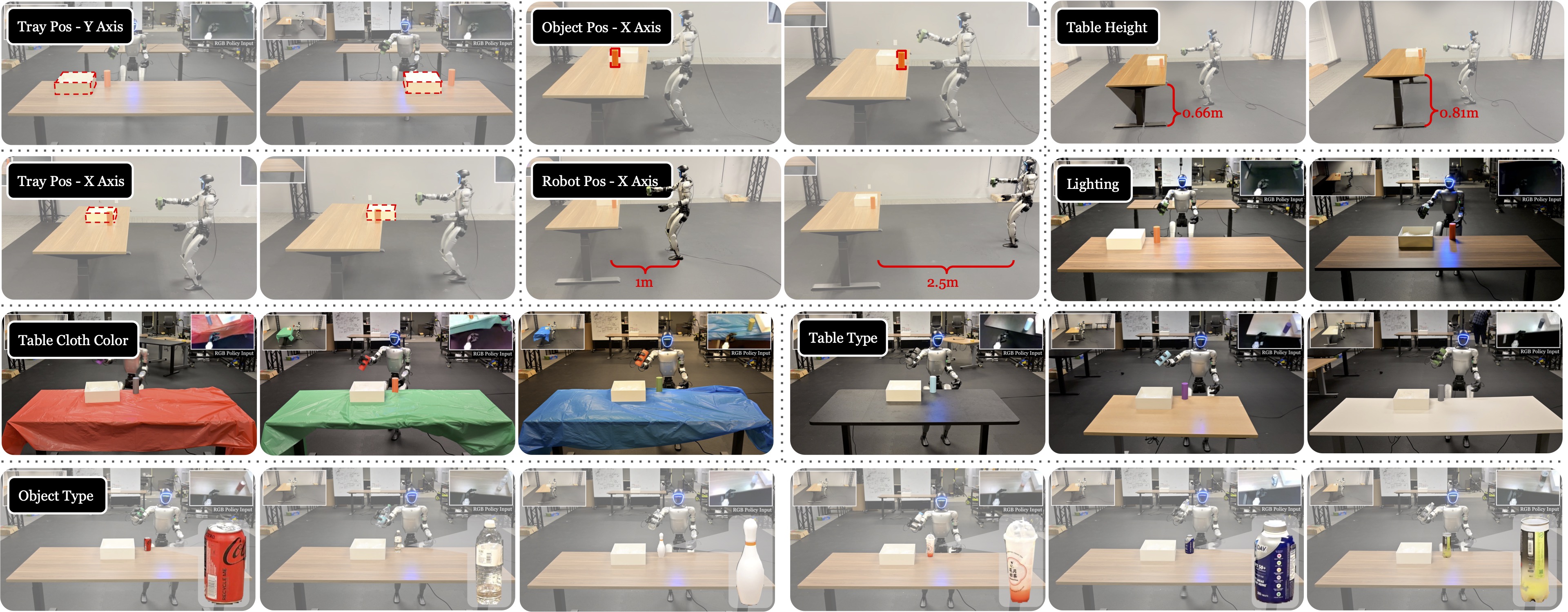}
    \vspace{-12pt}
   \caption{\textbf{Real-world generalization of \method RGB-based policy}
    under variations in tray and object position, robot start pose, table height and type, tablecloth color, lighting, and object category. Videos are provided in \href{https://viral-humanoid.github.io}{\text{https://viral-humanoid.github.io}}.}
    \vspace{-12pt}
   \label{fig:real_generalization}
\end{figure*}

\section{Real-World Results of \method}
\label{sec:real_world}

In this section, we present real-world humanoid loco-manipulation results achieved by \method. 
The following section (\Cref{sec:experiments}) analyzes the contribution of each design choice.
Our experiments deploy a 29-DoF Unitree G1 humanoid equipped with 7-DoF three-finger dexterous hands. 
Perception is provided by an Intel RealSense D435i, and all policy inference is performed on a desktop workstation with an Intel i9-14900K CPU and an NVIDIA RTX~4090 GPU.

\subsection{Robustness}
We evaluate the robustness of the learned student policy on a continuous loco-manipulation task in which the humanoid repeatedly walks between two tables, places an object, grasps a new object, and turns around. 
Across 59 consecutive real-world trials, \method succeeds in 54, demonstrating strong reliability under extended deployment.

We also compare \method with two human teleoperators: an expert with over 1000 hours of G1 teleoperation experience and a non-expert teleoperator with approximately one hour of experience. 
All conditions use the same HOMIE policy, yielding a near-apple-to-apple comparison. 
As shown in \Cref{fig:real_exp}, the expert attains a 100\% success rate with a 21.4\,s cycle time, slightly higher than the 20.2\,s cycle time of \method. 
Meanwhile, the non-expert reaches only 73\% success with significantly slower execution. These results show that although expert-level success remains challenging, \method achieves \textit{near–expert success performance while being faster than the expert}, and it substantially outperforms non-experts in both reliability and efficiency—highlighting its potential to reduce human workload in assisted teleoperation settings.

\subsection{Generalization}
We assess real-world generalization by systematically varying the environment along multiple factors, including tray start position, robot start pose, table height, lighting, table cloth, table type and color, and object category (\Cref{fig:real_generalization}). 
Across these variations, \method consistently completes the task without additional tuning, indicating strong robustness. 
We attribute this behavior to the domain randomization used during simulation training and the robustness of RL, which exposes the policy to diverse visual and spatial conditions. Videos are provided in \href{https://viral-humanoid.github.io}{\text{https://viral-humanoid.github.io}}.

\section{Experiments}
\label{sec:experiments}

In this section, we evaluate the contribution of each design component of \method, corresponding to the key elements introduced in \Cref{sec:methods}.

\subsection{Reference State Initialization for RL}
\begin{figure}[htbp]
  \centering
  \vspace{-5pt}
   \includegraphics[width=1.0\linewidth]{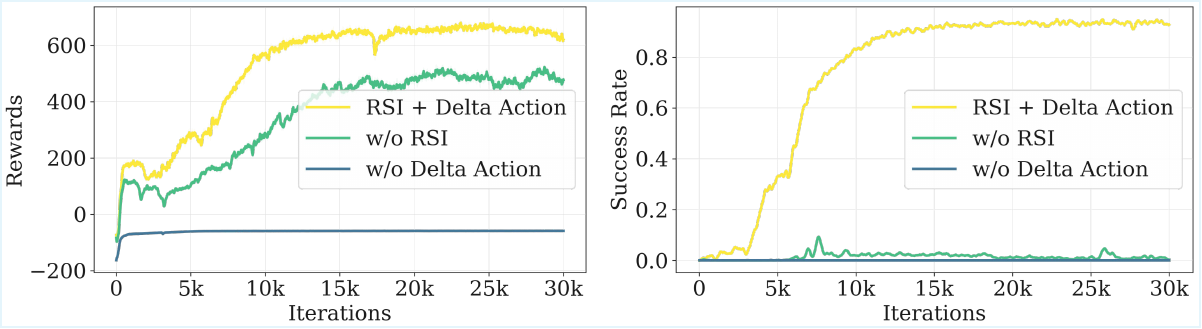}
    \vspace{-17pt}
   \caption{
   \textbf{Ablations of teacher policy training.}
   Training rewards (left) and success rates (right) for the full method (RSI + delta action), without demonstration resets, and without delta action space, showing that both components are critical for final success. 
   }
   \vspace{-12pt}
   \label{fig:TeacherResetDelta}
\end{figure}

Figure~\ref{fig:TeacherResetDelta} compares training curves with and without reference state initialization (RSI) \cite{peng2018deepmimic} from teleoperated demonstrations. 
Without RSI, the teacher policy quickly plateaus with a success rate below 10\%, whereas the full \method{} teacher with RSI reaches nearly 95\% success. 
RSI improves exploration by resetting episodes to diverse intermediate states along the task trajectory, so the policy can practice all stages of the task from the outset rather than discovering subgoals sequentially.

% \subsection{WBC-as-API v.s. Raw Motor Action Space}

\subsection{Delta versus Absolute Action Space}
We compare delta and absolute joint action spaces for the teacher policy. 
Unlike much of the legged locomotion RL literature, which commonly uses absolute joint targets, we find that a delta action space is crucial for humanoid loco-manipulation: as shown in Figure~\ref{fig:TeacherResetDelta}, only the delta-action teacher reliably solves the task, while the absolute-action variant fails to reach high success.

\subsection{Vision Backbone}

\begin{figure}[htbp]
  \centering
  \vspace{-2pt}
   \includegraphics[width=1.0\linewidth]{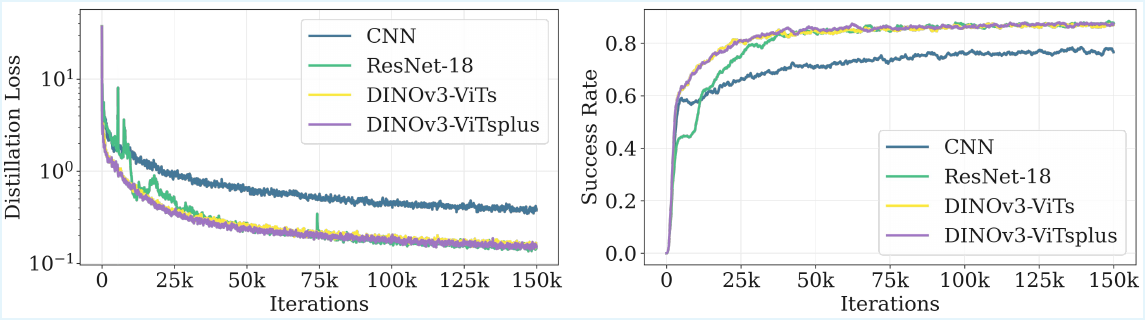}
    \vspace{-15pt}
   \caption{Ablation of vision backbone for student policy.}
   \label{fig:VisionBackbone}
   \vspace{-12pt}
\end{figure}

Figure~\ref{fig:VisionBackbone} reports the student’s training loss and success rate. We see that state-of-the-art vision backbones (DINOv3~\cite{{simeoni2025dinov3}}) yield stronger visual representations and greater capacity, enabling faster convergence and higher task success—i.e., the policy learns the target behaviors more reliably with better visual features.

\subsection{DAgger versus BC Visual Distillation}

\begin{figure}[htbp]
  \centering
  \vspace{-2pt}
   \includegraphics[width=1.0\linewidth]
   {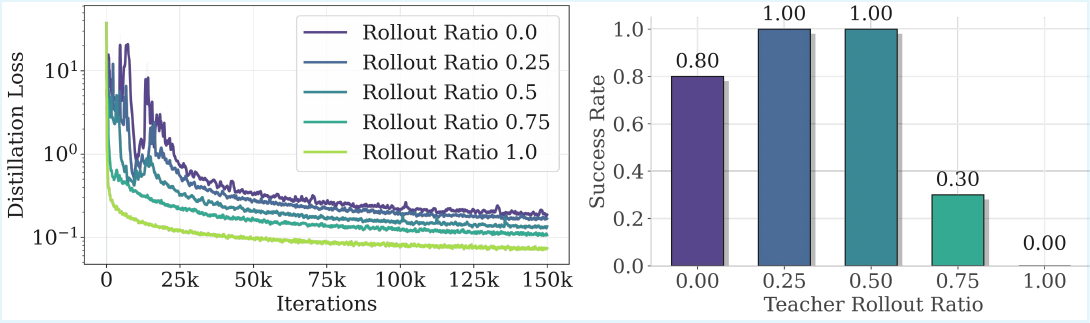}
   \vspace{-15pt}
   \caption{Ablation of ratio of DAgger/BC of student policy. }
    \vspace{-12pt}
   \label{fig:teacher_rollout_ratio}
\end{figure}

We ablate the DAgger-BC mixture by varying the rollout ratio $\alpha$, defined as the fraction of environments that follow the teacher policy during data collection (\(\alpha=0\) corresponds to pure DAgger on student rollouts, \(\alpha=1\) to pure BC). 
As shown in Figure~\ref{fig:teacher_rollout_ratio}, BC (\(\alpha=1\)) yields fast loss reduction but produces a brittle policy that fails to correct its own mistakes and performs poorly in Isaac-to-MuJoCo~\cite{mittal2025isaac,todorov2012mujoco} and real-world evaluations. 
Introducing student rollouts (\(\alpha=0.5\)) slows optimization slightly but substantially improves deployment success rate, so we adopt \(\alpha=0.5\) as our default DAgger-BC ratio.

\subsection{History Architecture}

\begin{figure}[htbp]
  \centering
  \vspace{-2pt}
   \includegraphics[width=1.0\linewidth]{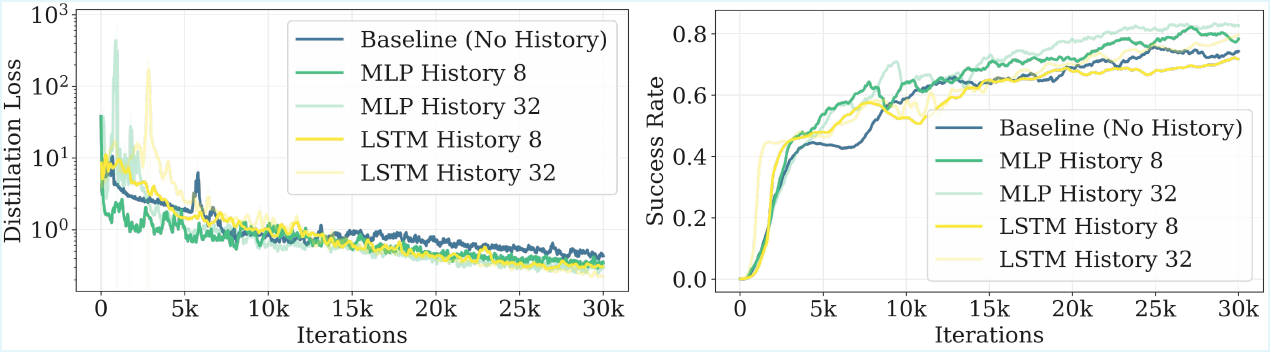}
   \vspace{-15pt}
   \caption{Ablation of the history architecture of student policy. }
    \vspace{-12pt}
   \label{fig:HistoryArch}
\end{figure}

Figure~\ref{fig:HistoryArch} compares a single-step baseline, a feed-forward history model, and an LSTM under different history lengths. 
History-aware models consistently outperform the single-step baseline, and longer temporal windows provide additional gains when resources allow.

\subsection{Visual Randomization}

\begin{wrapfigure}{r}{0.20\textwidth}
  % \vspace{-20pt}
  \hspace{-27pt} % reduce gap between text and figure
  \centering
  % \vspace{-25pt}
  \includegraphics[width=0.23\textwidth]{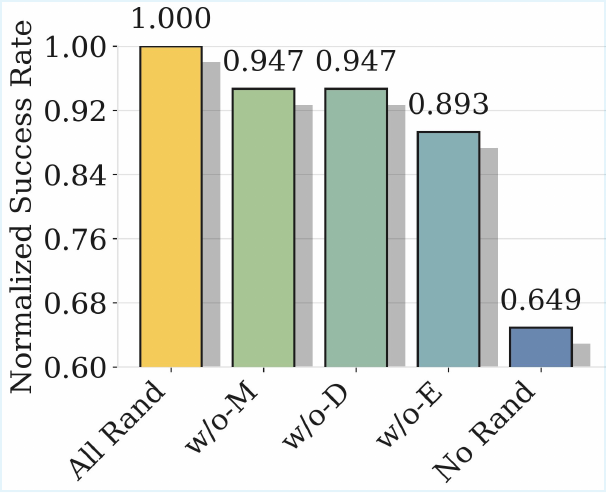}
  % \hspace{-10pt}
  \vspace{-10pt}
  \caption{Ablation of visual randomization.}
  \label{fig:VisualRandomizationAbaltion}
  \vspace{-8pt}
\end{wrapfigure}

Figure~\ref{fig:VisualRandomizationAbaltion} presents an ablation of our visual domain randomization. 
We focus on three dominant components: material randomization for table/floor/robot (M), dome-light randomization (D), 
and camera-extrinsics randomization (E), while other factors (image quality, object color, camera delay) provide smaller gains. 
All policies are evaluated in IsaacSim~\cite{mittal2025isaac} with \emph{all} randomizations enabled; training variants differ by removing one component at a time (w/o-M, w/o-D, w/o-E) or using no randomization at all. 
Success rates are normalized by the model trained with all randomizations (set to 1.0) and averaged over 200 episodes. 
Two trends emerge: (i) turning off all randomization causes a large drop in performance (down to 0.649, a 35.1\% decrease), and (ii) removing any single component also degrades performance, indicating that the randomizations are complementary and together form a crucial pipeline for robust sim-to-real transfer.

\subsection{Scaling Compute for Teacher Training}
\begin{figure}[htbp]
  \centering
    \vspace{-2pt}
   \includegraphics[width=1.0\linewidth]{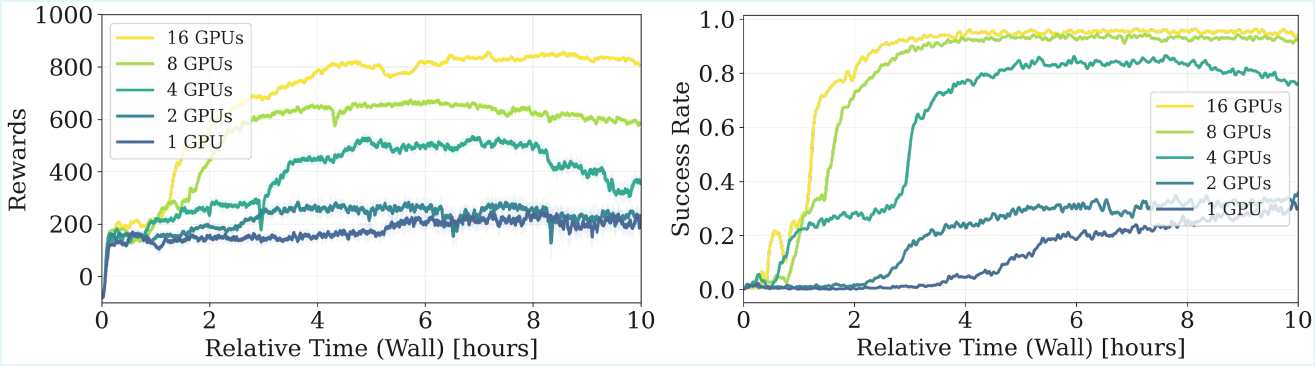}
    \vspace{-15pt}
    \caption{
    \textbf{Scaling compute for teacher training.}
    Rewards (left) and success rates (right) for 1–16 GPUs. 
    More GPUs yield faster convergence and better asymptotic performance.
    }
    \vspace{-12pt}
   \label{fig:TeacherScaling}
\end{figure}

Figure~\ref{fig:TeacherScaling} highlights the impact of scaling GPU resources from 1 to 16 during teacher training.
Increasing the number of GPUs substantially accelerates learning: larger batches of parallel environments broaden state-space coverage per unit wall time, enabling the policy to discover rewarding behaviors far more quickly.
Early training even shows better-than-linear speedup—for example, reaching a modest success rate of $\sim\!0.2$ with 4 GPUs takes well under half the time required with 2 GPUs—reflecting richer on-policy experience and more diverse rollouts.
Beyond speed, scaling has a pronounced effect on \emph{asymptotic performance}. 
With insufficient compute (1–2 GPUs), the teacher plateaus far below the desired performance range and never reaches high success rates.
In contrast, using 8–16 GPUs consistently drives the policy above 90\% success, revealing that large-scale simulation is not only beneficial but often \emph{necessary} for learning long-horizon humanoid loco-manipulation.

\subsection{Scaling Compute for Student Training}

\begin{figure}[htbp]
  \centering
   \includegraphics[width=1.0\linewidth]{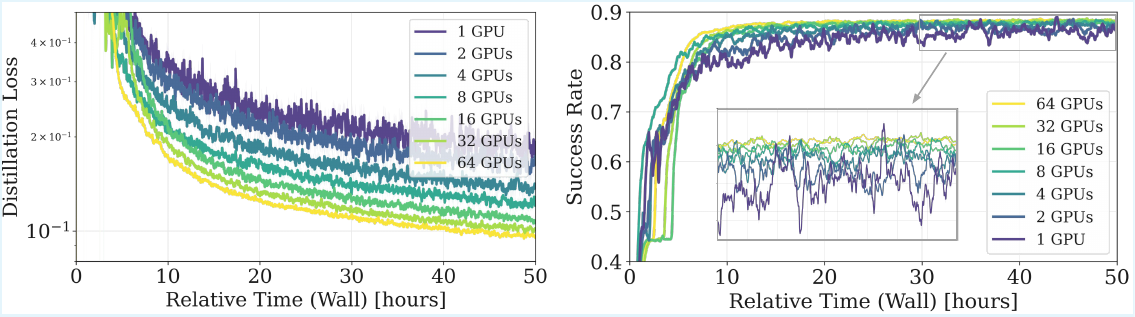}
   \vspace{-12pt}
    \caption{
    \textbf{Scaling compute for student policy training.} 
    Distillation loss (left) and success rate (right) when training with 1–64 GPUs. 
    Larger GPU counts provide significantly faster convergence, smoother optimization dynamics, and higher final performance, highlighting the importance of large-scale parallel simulation for vision-based loco-manipulation.
    }
    \vspace{-13pt}
   \label{fig:StudentScaling}
\end{figure}

We observe a clear scaling trend for the student policy as well. 
Figure~\ref{fig:StudentScaling} plots distillation (DAgger) loss and downstream success rate as we increase the number of GPUs from 1 to 64. 
Larger-scale training consistently accelerates convergence: the same loss threshold is reached dramatically sooner, and the success curve rises much more steeply. 
Beyond speed, scaling also improves training \emph{stability}: policies trained with more GPUs exhibit smoother loss curves and less variance in success rate, especially during the early stages when the student is most sensitive to distribution shift. 
Interestingly, higher-GPU runs also achieve slightly higher final success, suggesting that large-scale experience collection yields richer and more diverse state coverage, which in turn improves robustness. 
Overall, these results indicate that substantial computing is not merely a convenience but a practical requirement for reliable visual loco-manipulation distillation.

\subsection{Object generalization}

\begin{figure}[htbp]
  \centering
   \includegraphics[width=0.9\linewidth]{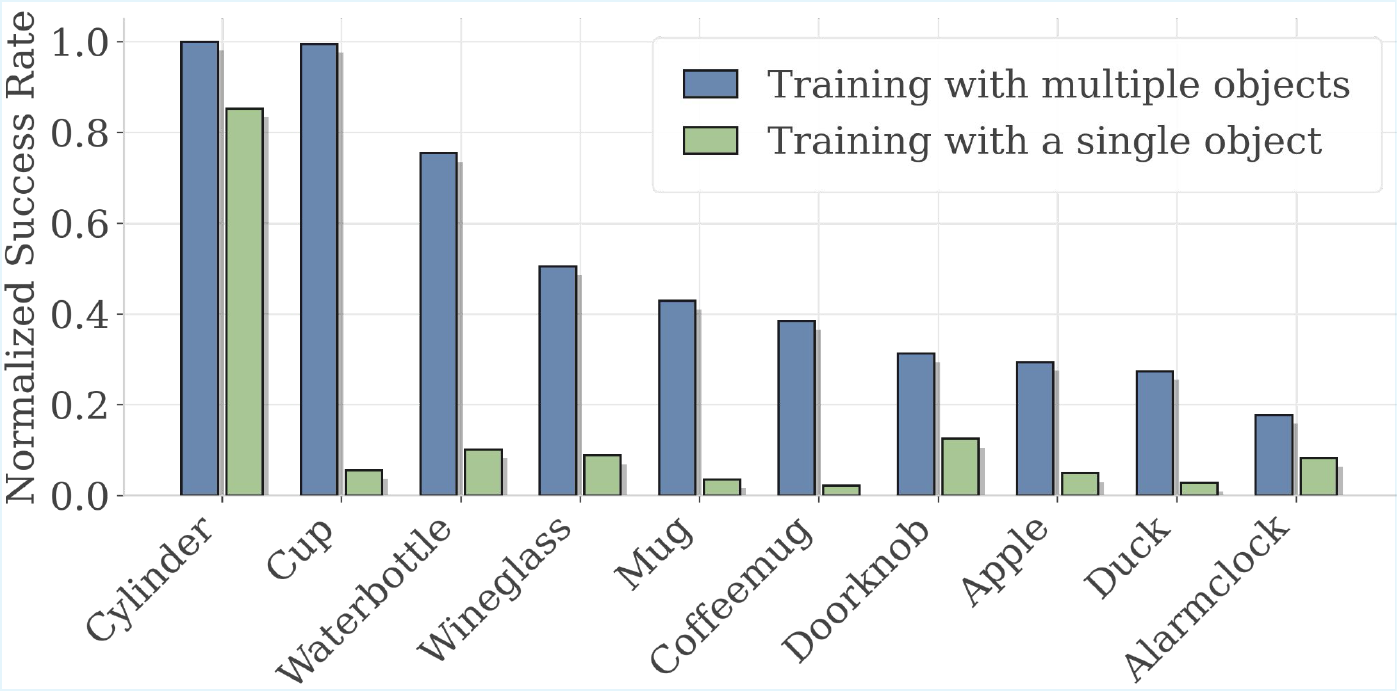}
   \vspace{-7pt}
   \caption{Ablation of object generalization of teacher policy.}
   \vspace{-18pt}
   \label{fig:ObjectGeneralization}
\end{figure}

We study object-level generalization on the grasping subtask under two training regimes: (i) single-object training on a cylinder only and (ii) multi-object training on ten distinct objects. 
At test time, we evaluate on the same ten objects and report normalized success rates.
As shown in Figure~\ref{fig:ObjectGeneralization}, training with multiple objects yields substantially better generalization—the multi-object policy attains higher success on every category than the cylinder-only baseline.

\vspace{-5pt}

\section{Related Work}
\label{sec:relatedworks}

\paragraph{Sim-to-Real for Locomotion}  
Sim-to-real techniques have enabled blind locomotion policies trained in simulation to be deployed zero-shot on real legged robots~\cite{tan2018sim,hwangbo2019learning,lee2020learning,kumar2021rma,liao2025berkeley,duan2022learning,ben2025homie,gu2024advancing,chen2024learning,li2025reinforcement,li2021reinforcement}. These proprioceptive policies are robust and agile but lack environmental awareness, making them insufficient for navigation in cluttered or goal-directed settings. To compensate, some works incorporate depth image or LiDAR-based elevation map to model terrain geometry~\cite{miki2022learning,zhuang2024humanoid,cheng2023parkour,wang2025beamdojo,long2025learning,ren2025vb}, improving foot placement but offering limited semantic understanding. Some works~\cite{cheng2024navila,wei2025streamvln,cai2025navdp} combine RGB vision and language instructions with sim-to-real locomotion policies for semantic navigation, but rely on high-latency vision-language-action (VLA) models. \method instead distills compact, RGB-driven visuomotor policies trained in randomized simulation to support real-time, goal-conditioned locomotion without sacrificing sim-to-real scalability.

\paragraph{Sim-to-Real for Manipulation}  
Visual sim-to-real has been a key driver of progress in manipulation. A central tool is domain randomization: by varying rendering properties in simulation, policies trained on RGB can transfer to the real world~\cite{sadeghi2016cad2rl,tobin2017domain,peng2018sim}. This strategy enabled end-to-end RL for challenging skills—e.g., OpenAI’s Dactyl reoriented objects with a five-finger hand using massive randomization and curriculum learning~\cite{andrychowicz2020learning,akkaya2019solving}, and later works scaled to tasks like Rubik’s Cube solving and high-speed rotation~\cite{akkaya2019solving,handa2023dextreme}. Early attempts that relied on high-fidelity RGB/depth/point-cloud simulation still struggled with the reality gap~\cite{yuan2024learning,jiang2024transic,ze2023visual,hansen2021generalization,huang2022spectrum,akkaya2019solving,andrychowicz2020learning}. Recent methods improve robustness and scalability via a teacher–student paradigm: a privileged-state teacher is trained first, then a student is distilled from RGB with randomization~\cite{singh2024dextrah,deng2025graspvla,liu2024visual}. However, prior work largely targets tabletop settings. We extend this paradigm to humanoid loco-manipulation.

\paragraph{Sim-to-Real for Loco-Manipulation}  
Loco-manipulation requires a humanoid robot to move through an environment while simultaneously interacting with objects. It poses unique challenges for sim-to-real learning. Recent work explores sim-to-real learning for low-level loco-manipulation control, using either modular architectures that decouple leg and arm~\cite{liu2024visual,ben2025homie,cheng2024express} or end-to-end policies that coordinate full-body motion~\cite{ha2024umi,pan2025roboduet,he2024omnih2o}. On top of these controllers, some systems achieve task-level loco-manipulation via imitation learning or vision–language–action models~\cite{ha2024umi,qiu2025wildlma,wang2025odyssey}, but these approaches require large real-world datasets and often lack robustness. \method bridges both layers by training an RGB-driven, end-to-end policy entirely in simulation, enabling zero-shot deployment for humanoid goal-conditioned loco-manipulation without real-world demonstrations or large models.

\section{Limitations and Discussions}
\label{sec:discussion}

While \textit{sim-to-real} has demonstrated remarkable success in isolated capabilities—robust locomotion, geometric perception, and rigid-body manipulation—scaling these methods to general-purpose loco-manipulation (\textit{``locomote anywhere, perceive anything, manipulate everything''}) exposes four critical coverage gaps that current paradigms have yet to bridge.

\paragraph{\texttt{Physics Coverage}:\\ \textbf{The Physical Diversity Gap}}
Modern simulators theoretically possess the capability to model complex dynamics, including fluid-structure interactions and deformable bodies. The fundamental bottleneck is not the lack of simulation features, but the \textit{scalability of engineering effort} required to ground these features in reality. We can, with sufficient effort, engineer specific environments to simulate scooping rice, grasping noodles with tongs, cutting garlic, hand-crafting sushi, or feeding beans into a coffee machine. However, each of these scenarios requires bespoke tuning of material properties and boundary conditions to align with the real world. The challenge lies in scaling this effort to the open-ended diversity of daily life: modeling the damping of every cardboard box, the stiffness of every garment, the friction of specific oil stains, or the granular mechanics of food items. The barrier is not that we \textit{cannot} simulate these interactions, but that the engineering cost to accurately instantiate them for the long tail of real-world physics arguably exceeds the complexity of collecting real-world data itself.

\vspace{-2pt}
\paragraph{\texttt{Task Coverage}:\\ \textbf{The Long-Tail of Task Generation}}
Even if physics could be perfectly simulated, the diversity of tasks remains an unresolved challenge. Constructing a simulation environment for a single task (e.g., dishwashing) requires modeling not just object geometries, but their functional affordances, varied states (dirty vs. clean), and interaction logic. Scaling this to the thousands of distinct chores in a household environment presents a massive content generation bottleneck. 
Furthermore, simulation is limited by human imagination; we cannot simulate ``unknown unknowns''—edge cases and task variants that only emerge during real-world deployment (e.g., adapting to a pet's interference or accommodating a human user with mobility constraints). Current asset taxonomies and generative procedural pipelines fail to capture this functional breadth.

\vspace{-2pt}
\paragraph{\texttt{Reward and Policy Coverage}:\\ \textbf{The Reward Engineering Bottleneck}}
Defining ``RL-friendly'' reward functions that are both dense enough to guide exploration and sparse enough to prevent specification gaming is a delicate art that does not scale. In practice, we observe a tension between \textit{under-exploration} (where dense, shaped rewards bias the policy toward local optima or simulator exploits) and \textit{over-exploration} (where sparse rewards fail to bootstrap learning in high-dimensional spaces). 
For a single task, tuning these rewards to find the ``Goldilocks'' regime is feasible. However, manually designing robust reward functions for thousands of distinct tasks is intractable. This highlights a crucial trade-off: while sim-to-real offers scalable data generation, it demands high upfront engineering effort. In contrast, imitation learning moves the burden to data collection. As it stands, a few days of high-quality teleoperation data can often outperform months of sim-to-real engineering for specific tasks, primarily because the ``reward'' is implicitly provided by the human demonstrator, bypassing the specification problem entirely.

\vspace{-2pt}
% \paragraph{\textbf{Hardware Coverage: The Hardware-Simulation Gap}}
\paragraph{\texttt{Hardware Coverage}:\\ \textbf{The Hardware-Simulation Gap}}
Finally, a distinct gap remains between the idealized actuation in simulation and the reality of current humanoid hardware. While quasi-direct drive (QDD) actuators for locomotion are relatively well-modeled, dexterous manipulation hardware often suffers from unmodeled friction, backlash, thermal throttling, and sensor noise. Simulation policies that rely on precise finger positioning or force feedback often fail to transfer to hardware that lacks the requisite reliability and precision, limiting the complexity of tasks that can be genuinely attempted in the real world.

\vspace{-2pt}
\paragraph{\textbf{Outlook}}
These four gaps suggest that while sim-to-real will retain a critical role in robotics—particularly for safe, stable evaluation and solving skills with bounded state-spaces—scaling it to solve general-purpose loco-manipulation is likely out of reach for the near future. The field has successfully identified the \textit{sweet spot} for sim-to-real in locomotion: where aggressive randomization of limited parameters (terrain, mass) and carefully-designed reward functions produce robust policies that generalize well. However, the equivalent sweet spot for manipulation remains undiscovered, as the complexity of contact physics and semantic diversity in manipulation vastly exceeds that of locomotion tasks.

We believe the path forward involves redefining the role of simulation within a broader data ecosystem. Rather than forcing simulation to generate the entire distribution of the real world, the next frontier lies in integrating sim-to-real with the rapidly maturing stacks of real-world imitation learning and foundation models. 
Discovering this synergy—where simulation complements rather than replaces real-world learning—is the most exciting direction for the future of general-purpose loco-manipulation.

% Comment for submission
% \input{sec/n_discussion}
\section*{Acknowledgement}
We thank Jeremy Chimienti, Tri Cao, Jazmin Sanchez, Isabel Zuluaga, Jesse Yang, Caleb Geballe, Haotian Lin, Lingyun Xu, Huanyu Li, Chaoyi Pan, Chaitanya Chawla, Jason Liu, Tony Tao, Ritvik Singh, Ankur Handa, Arthur Allshire, Guanzhi Wang, Yinzhen Xu, Runyu Ding, Xiaowei Jiang, Yuqi Xie, Jimmy Wu, Haoyu Xiong, Avnish Narayan, Kaushil Kundalia, Qi Wang, Scott Reed, Ziang Cao, Fengyuan Hu, Sirui Chen, Chenran Li, and Tingwu Wang for their help and support during this project.

\clearpage
{
    \small
    \bibliographystyle{ieeenat_fullname}
    \bibliography{main}
}

\clearpage
\setcounter{page}{1}
\maketitlesupplementary

\section{Training Details}
\label{appendix:training}

\subsection{Observation Details}
\label{appendix:observation}

Table~\ref{tab:teacher-obs} summarizes the observation terms and their corresponding dimensions.

\begin{table}[h]
\centering
\small
\begin{tabular}{l c}
\toprule
\textbf{State term} & \textbf{Dimensions} \\
\midrule
Base linear velocity              & 3 \\
Base angular velocity             & 3 \\
Projected gravity                 & 3 \\
Actions                           & 31 \\
Stage                             & 5 \\
Delta actions                     & 11 \\
DoF position                      & 43 \\
DoF velocity                      & 43 \\
Placement position                & 2 \\
Table--pelvis transform           & 9 \\
Finger-tip forces for hold\_object            & 12 \\
Hold\_object transform             & 9 \\
Hold\_object--hand transform       & 9 \\
Target pre-place position             & 3 \\
Finger-tip forces for grasp\_object          & 12 \\
Grasp\_object transform            & 9 \\
Grasp\_object--hand transform      & 9 \\
Target lift position              & 3 \\
HOMIE commands & 7 \\
\midrule
Single-step total dim             & 226 \\
\bottomrule
\end{tabular}
\caption{Observation dimensions for \texttt{teacher}.}
\label{tab:teacher-obs}
\end{table}

Table~\ref{tab:student-obs} lists the observation terms and their corresponding dimensions. In addition to these state observations, we feed an RGB image of size $108\times192$ into the vision encoder. The resulting 128-dimensional visual feature is concatenated with the state observations and then passed to the policy head.

\begin{table}[h]
\centering
\small
\begin{tabular}{l c}
\toprule
\textbf{State term} & \textbf{Dimensions} \\
\midrule
Base angular velocity             & 3 \\
Projected gravity                 & 3 \\
Actions                           & 31 \\
DoF position (w/o fingers)        & 29 \\
DoF velocity (w/o fingers)        & 29 \\
Delta actions                     & 11 \\
HOMIE commands  & 7 \\
\midrule
Single-step total dim             & 113 \\
\bottomrule
\end{tabular}
\caption{Observation dimensions for \texttt{student}.}
\label{tab:student-obs}
\end{table}

\subsection{Reward Details}
\label{appendix:reward}

A single place–pickup cycle is decomposed into five stages: (1) walking toward the object; (2) moving the arm and hand to a pre-place pose; (3) placing the object; (4) grasping and lifting the next object; and (5) turning. Repeating this sequence produces a long-horizon loco-manipulation loop. At each step, the total reward is a stage-weighted sum
\begin{equation*}
r_t = \sum_{i=0}^{4} w_i \mathbbm{1}[s_t = i] r^{(i)}_t,\qquad w_i > 0,
\end{equation*}
and stage transitions are governed by stage-specific advancement and completion criteria. Table~\ref{tab:wsdpt-rewards-stages} instantiates $r^{(s)}$ with stage-dependent shaping terms for teacher policy.

\begin{table*}[t]
\centering
\footnotesize
\setlength{\tabcolsep}{3pt}
\renewcommand{\arraystretch}{1.1}
\begin{tabular}{l l r c}
\toprule
\textbf{Term} & \textbf{Expression} & \textbf{Weight} & \textbf{Stage(s)} \\
\midrule
\multicolumn{4}{l}{\textbf{Termination / generic penalties}} \\
Termination
  & $\mathbbm{1}_{\{\text{termination}\}}$
  & $-2000.0$ & $0$--$4$ \\

Delta action rate
  & $\|\Delta a_t\|_2^{2}$
  & $-0.01$ & $0$--$4$ \\

DoF velocity
  & $\|\dot{\mathbf{q}}\|_2^{2}$
  & $-0.5$ & $0$--$4$ \\

DoF acceleration
  & $\|\ddot{\mathbf{q}}\|_2^{2}$
  & $-3.0\times 10^{-6}$ & $0$--$4$ \\

Torque limits
  & $\|\boldsymbol{\tau}\|_2^{2}$
  & $-0.001$ & $0$--$4$ \\

Output smoothness
  & $\|\pi_t - \pi_{t-1}\|_2^{2}$
  & $-9.0$ & $0$--$4$ \\

Finger primitive limits
  & $\bigl|\operatorname{clip}(u_{\text{finger}},[l,u]) - u_{\text{finger}}\bigr|$
  & $-20.0$ & $0$--$4$ \\

Fast right-arm velocity
  & $\|\dot{\mathbf{q}}_{\text{right arm}}\|_2^{2}$
  & $-80.0$ & $0$--$4$ \\

Finger qvel, when contacting ground with single-foot
  & $\|\dot{\mathbf{q}}_{\text{finger}}\|_2\,\mathbbm{1}_{\text{single-foot}}$
  & $-3000.0$ & $1$--$3$ \\

Arm qvel, when contacting ground with single-foot
  & $\|\dot{\mathbf{q}}_{\text{right arm}}\|_2\,\mathbbm{1}_{\text{single-foot}}$
  & $-1300.0$ & $1$--$3$ \\

\midrule
\multicolumn{4}{l}{\textbf{Heading / command shaping}} \\

Heading toward object
  & $\bigl((\psi_{\text{GraspObj}}-\psi_{\text{robot}})/\pi\bigr)^{2}$
  & $-10000.0$ & $0$ \\

Object in view
  & $\mathbbm{1}[y_{\text{right hand}} > y_{\text{GraspObj}}-0.1]+\mathbbm{1}[y_{\text{left hand}} < y_{\text{GraspObj}}+0.1]$
  & $-1.0$ & $0$ \\

Large linear $v_x$ command
  & $\sum\max\bigl(0, |v_x^{\text{cmd}}|-0.5\bigr)$
  & $-20.0$ & $0$--$4$ \\

Large linear $v_y$ command
  & $\sum\max\bigl(0, |v_y^{\text{cmd}}|-0.5\bigr)$
  & $-20.0$ & $0$--$4$ \\

Large angular $\omega$ command
  & $\sum\max\bigl(0, |\omega^{\text{cmd}}|-0.5\bigr)$
  & $-20.0$ & $0$--$4$ \\

Large upper-body actions
  & $\sum\max\bigl(0, |u_{\text{upper}}|-2\pi\bigr)$
  & $-20.0$ & $0$--$4$ \\

Zero linear $v_x$, linear $v_y$, angular $\omega$ cmd
  & $|v_x^{\text{cmd}}| + |v_y^{\text{cmd}}| + |\omega^{\text{cmd}}| $
  & $-12.0$ & $1$--$3$ \\

Zero linear $v_x$, linear $v_y$ cmd
  & $|v_x^{\text{cmd}}| + |v_y^{\text{cmd}}|$
  & $-4.0$ & $4$ \\

\midrule
\multicolumn{4}{l}{\textbf{Task / object-centric rewards}} \\

Robot-Object distance
  & $\exp(-4\,(\|p_\text{robot} - p_\text{GraspObj}\| - 0.45)^2)$
  & $2.0$ & $0$--$4$ \\

Upper-body actions (pose)
  & $\|\mathbf{q}_{\text{right arm}}\|_2^{2}$
  & $-1.0$ & $0$ \\

Keep hand closed
  & $\exp \bigl(-4\,(u_{\text{finger}}-u_{\text{close}})^{2}\bigr)$
  & $9.0$ & $0$--$1$, $3$--$4$ \\

Place objects when near tray
  & $- \|\mathbf f_{\text{PlaceObj}}\| * \mathbbm{1} (\|p_\text{PlaceObj} - p_\text{tray}\| < 0.3)$
  & $10.0$ & $0$--$1$ \\

Holding object
  & $\exp \bigl(-4\|p_{\text{PlaceObj}}-p_{\text{hand}}\|_2\bigr)$
  & $1.0$ & $0$--$4$ \\

Hand–object distance
  & $\exp \bigl(-10\max_k\|p^{(k)}_{\text{finger}}-p_{\text{GraspObj}}\|_2\bigr)$
  & $20.0$ & $3$--$4$ \\

Grasp based on obj–finger dir
  & $-\;\hat{\mathbf{d}}_{\text{thumb}}^{\top}\hat{\mathbf{d}}_{\text{index}}$
  & $5.0$ & $3$--$4$ \\

Grasp force
  & $\sum\|\mathbf{f}_{\text{GraspObj-hand}}\|$
  & $1.0$ & $3$--$4$ \\

Lift goal distance
  & $\exp(-10||p_\text{GraspObj} - p_\text{goal}||^2)$
  & $10.0$ & $3$--$4$ \\

Lift $z$
  & $\mathrm{min}(h_\text{GraspObj} - h_\text{table}, 0.15)$
  & $200.0$ & $3$--$4$ \\

Turn around
  & $-|\text{y}_\text{robot} - \text{y}_\text{desired}|$
  & $15.0$ & $4$ \\

Right-arm qpos tracking (hold)
  & $\exp \bigl(-4\|\mathbf{q}_{\text{right arm}}-\mathbf{q}^{*}_{\text{Place}}\|_2\bigr)$
  & $5.0$ & $0$--$2$ \\

Right-arm qpos tracking (front)
  & $\exp \bigl(-4\|\mathbf{q}_{\text{right arm}}-\mathbf{q}^{*}_{\text{Grasp}}\|_2\bigr)$
  & $25.0$ & $3$--$4$ \\

Finger qvel during right-arm qvel
  & $\exp \bigl(-6\|\dot{\mathbf{q}}_{\text{arm}}\|_2
                    \|\dot{\mathbf{q}}_{\text{finger}}\|_2\bigr)$
  & $15.0$ & $1$--$4$ \\

Object–table contact move
  & $\|\mathbf{v}_{\text{GraspObj},xy}\|\;\mathbbm{1}_{\text{table-contact}}$
  & $-1000.0$ & $1$--$4$ \\

Object relative move (hand–obj $v_z$)
  & $|v^{z}_{\text{GraspObj}}-v^{z}_{\text{hand}}|\;\mathbbm{1}_{\text{in-grasp}}$
  & $-3000.0$ & $1$--$3$ \\

Object lean during pick
  & $|\phi_{\text{GraspObj}}|+|\theta_{\text{GraspObj}}|$
  & $-500.0$ & $0$--$3$ \\

Object non–$z$ velocity during pick
  & $\|\mathbf{v}_{\text{GraspObj},xy}\|_2$
  & $-500.0$ & $0$--$3$ \\
\bottomrule
\end{tabular}
\caption{Reward components, expressions, weights, and the stages ($0$--$4$) where each term is applied.}
\label{tab:wsdpt-rewards-stages}
\end{table*}

\subsection{Hyperparameters Details}
\label{appendix:hyperparameters}

Table~\ref{tab:pd} lists the PD gains used for the Unitree G1 robot equipped with 3-finger dexterous hands.

Table~\ref{tab:hyperparams-teacher} lists the hyperparamters for teacher policy trained by PPO~\cite{schulman2017proximal}.

Table~\ref{tab:hyperparams-student} lists the hyperparameters for student policy trained by the mixture of DAgger~\cite{ross2011dagger} and Behavior Cloning.

\begin{table}[t]
\centering
\small
\begin{tabular}{lcc}
\toprule
\textbf{Joint} & \textbf{$K_p$ [N$\cdot$m/rad]} & \textbf{$K_d$ [N$\cdot$m$\cdot$s/rad]} \\
\midrule
hip\_yaw        & 150   & 2.0 \\
hip\_roll       & 150   & 2.0 \\
hip\_pitch      & 150   & 2.0 \\
knee            & 200   & 4.0 \\
ankle\_pitch    & 40    & 2.0 \\
ankle\_roll     & 40    & 2.0 \\
waist\_yaw      & 250   & 5.0 \\
waist\_roll     & 250   & 5.0 \\
waist\_pitch    & 250   & 5.0 \\
shoulder\_pitch & 100   & 5.0 \\
shoulder\_roll  & 100   & 5.0 \\
shoulder\_yaw   & 40    & 2.0 \\
elbow           & 40    & 2.0 \\
wrist\_roll     & 20    & 2.0 \\
wrist\_pitch    & 20    & 2.0 \\
wrist\_yaw      & 20    & 2.0 \\
hand\_index     & 0.5   & 0.1 \\
hand\_middle    & 0.5   & 0.1 \\
hand\_thumb\_1  & 0.5   & 0.1 \\
hand\_thumb\_2  & 0.5   & 0.1 \\
hand\_thumb\_0  & 2.0   & 0.1 \\
\bottomrule
\end{tabular}
\caption{Joint-space PD gains ($K_p$, $K_d$) used in the low-level controller.}
\label{tab:pd}
\end{table}

\begin{table}[t]
\centering
\small
\begin{tabular}{l c}
\toprule
\textbf{Hyperparameters} & \textbf{Values} \\
\midrule
Number of environments                 & 32768 (2048*8GPUs*2Nodes) \\
Discount factor ($\gamma$) & 0.998 \\
Learning rate              & 0.00002 \\
Entropy coefficient               & 0.01 \\
Value loss coefficient            & 1 \\
Init noise std (RL)        & 0.5 \\
MLP size                   & [512, 256, 128] \\
\bottomrule
\end{tabular}
\caption{Hyperparameters for teacher policy.}
\label{tab:hyperparams-teacher}
\end{table}

\begin{table}[t]
\centering
\footnotesize
\begin{tabular}{l c}
\toprule
\textbf{Hyperparameters} & \textbf{Values} \\
\midrule
Number of environments                 & 65535 (1024*8GPUs*8Nodes) \\
Number of steps per environment & 1 \\
Learning rate              & 0.0002 \\
\bottomrule
\end{tabular}
\caption{Hyperparameters for student policy.}
\label{tab:hyperparams-student}
\end{table}

\subsection{Domain Randomization}

Table~\ref{tab:randomization} summarizes all randomizations used during policy training, including image quality, dome lighting, materials, table properties, and camera extrinsics.

\begin{table*}[ht]
\centering
\small
\caption{Comprehensive domain randomization parameters during training}
\begin{tabular}{lcc}
\toprule
\textbf{Parameter} & \textbf{Probability} & \textbf{Distribution} \\
\midrule
\multicolumn{3}{l}{\textbf{Image Augmentation}} \\
Brightness & 0.25 & $\sim \mathcal{U}(0.7, 2)$ \\
Contrast & 0.25 & $\sim \mathcal{U}(0.5, 1.5)$ \\
Hue & 0.5 & $\sim \mathcal{U}(-0.1, 0.1)$ \\
Saturation & 0.25 & $\sim \mathcal{U}(0.5, 2)$ \\
Gaussian Noise Std & 0.25 & $\sim \mathcal{U}(0.0, 0.15)$ \\
Gaussian Blur Kernel Size & 0.25 & $\sim \mathcal{U}(3, 5)$ \\
Gaussian Blur Sigma & 0.25 & $\sim \mathcal{U}(0.1, 1.5)$ \\
\midrule
\multicolumn{3}{l}{\textbf{Lighting}} \\
Dome Light Intensity & 1.0 & $\sim \mathcal{U}(800, 2000)$ \\
Dome Light Yaw Rotation & 1.0 & $\sim \mathcal{U}(-\pi, \pi)$ \\
Dome Light Texture Map & 1.0 & $\sim \mathcal{U}(\text{texture\_maps})$ \\
 & & (Indoor, Clear, Cloudy, Night, Studio) \\
\midrule
\multicolumn{3}{l}{\textbf{Material Randomization}} \\
Robot Material - Roughness & 1.0 & $\sim \mathcal{U}(0.0, 0.8)$ \\
Robot Material - Metallic & 1.0 & $\sim \mathcal{U}(0.0, 0.8)$ \\
Robot Material - Specular & 1.0 & $\sim \mathcal{U}(0.0, 0.8)$ \\
Floor Material Texture & 1.0 & $\sim \mathcal{U}(\text{texture\_maps})$ \\
 & & (Wood, Carpet, Masonry, Metals, \\
 & & Natural, Plastics, Stone, Wall Board) \\
Table Material Texture & 1.0 & $\sim \mathcal{U}(\text{texture\_maps})$ \\  
& & (Wood) \\
Object Material Texture & 1.0 & $\sim \mathcal{U}(\text{texture\_maps})$ \\
 & & (All Base Materials) \\
\midrule
\multicolumn{3}{l}{\textbf{Table Physical Properties}} \\
Table Height (m) & 1.0 & $\sim \mathcal{U}(0.65, 0.6775)$ \\
Table Depth (m) & 1.0 & $\sim \mathcal{U}(0.7, 0.75)$ \\
Table Width (m) & 1.0 & $\sim \mathcal{U}(1.4, 1.6)$ \\
Table Thickness (m) & 1.0 & $\sim \mathcal{U}(0.035, 0.04)$ \\
\midrule
\multicolumn{3}{l}{\textbf{Camera Extrinsics}} \\
Position Noise - X (m) & 1.0 & $\sim \mathcal{U}(-0.02, 0.02)$ \\
Position Noise - Y (m) & 1.0 & $\sim \mathcal{U}(-0.05, 0.05)$ \\
Position Noise - Z (m) & 1.0 & $\sim \mathcal{U}(-0.02, 0.02)$ \\
Rotation Noise - Roll (rad) & 1.0 & $\sim \mathcal{U}(-0.05, 0.05)$ \\
Rotation Noise - Pitch (rad) & 1.0 & $\sim \mathcal{U}(-0.1, 0.1)$ \\
Rotation Noise - Yaw (rad) & 1.0 & $\sim \mathcal{U}(-0.05, 0.05)$ \\
\bottomrule
\end{tabular}
\label{tab:randomization}
\end{table*}

\end{document}